\documentclass[runningheads]{llncs}
\usepackage{graphicx}
\usepackage{appendix}
\usepackage{biblatex} 
\usepackage{booktabs,multirow,longtable,makecell} 
\usepackage{subfigure}
\begin{document}
\title{WeLayout: WeChat Layout Analysis System for the ICDAR 2023 Competition on Robust Layout Segmentation in Corporate Documents}
%
%

\author{Mingliang Zhang\textsuperscript{$\star$} \and
Zhen Cao\textsuperscript{$\star$} \and
Juntao Liu \and
Liqiang Niu\textsuperscript{$\dagger$} \and
Fandong Meng \and
Jie Zhou}
%
\titlerunning{ }
\authorrunning{ }
%
\institute{Pattern Recognition Center, WeChat AI, Tencent Inc, China
\email{\{mmlzhang,zhenzcao,gentoliu,poetniu,fandongmeng,withtomzhou\}@tencent.com}}
%
\maketitle              

\renewcommand{\thefootnote}{\fnsymbol{footnote}}
\footnotetext[1]{Equal contributions.}
\footnotetext[4]{Corresponding author.}
\renewcommand{\thefootnote}{\arabic{footnote}}

\begin{abstract}
In this paper, we introduce \textbf{WeLayout}, a novel system for segmenting the layout of corporate documents, which stands for \textbf{We}Chat \textbf{Layout} Analysis System. Our approach utilizes a sophisticated ensemble of DINO and YOLO models, specifically developed for the ICDAR 2023 Competition on Robust Layout Segmentation. Our method significantly surpasses the baseline, securing a top position\footnote{https://eval.ai/web/challenges/challenge-page/1923/leaderboard/4545} on the leaderboard with a mAP of 70.0. To achieve this performance, we concentrated on enhancing various aspects of the task, such as dataset augmentation, model architecture, bounding box refinement, and model ensemble techniques. Additionally, we trained the data separately for each document category to ensure a higher mean submission score. We also developed an algorithm for cell matching to further improve our performance. To identify the optimal weights and IoU thresholds for our model ensemble, we employed a Bayesian optimization algorithm called the Tree-Structured Parzen Estimator. Our approach effectively demonstrates the benefits of combining query-based and anchor-free models for achieving robust layout segmentation in corporate documents.

\end{abstract}

\section{Introduction}
Corporate documents, such as financial reports, invoices, and contracts, contain valuable information that must be extracted and analyzed. However, the intricate and diverse layouts of these documents present challenges for automated systems in accurately segmenting the content. Recognizing this challenge, the International Conference on Document Analysis and Recognition organized a competition on robust layout segmentation in corporate documents for 2023\footnote{https://ds4sd.github.io/icdar23-doclaynet}.

The competition's objective is to promote the development of precise and efficient algorithms for document layout segmentation. It provides participants with a dataset of diverse corporate documents, including financial reports, invoices, and contracts, featuring various layouts and structures. Participants are required to develop algorithms that can accurately segment the content of these documents into logical units, such as paragraphs, tables, and figures.

\begin{figure}[!h] 
\centering 
\includegraphics[width=0.65\textwidth]{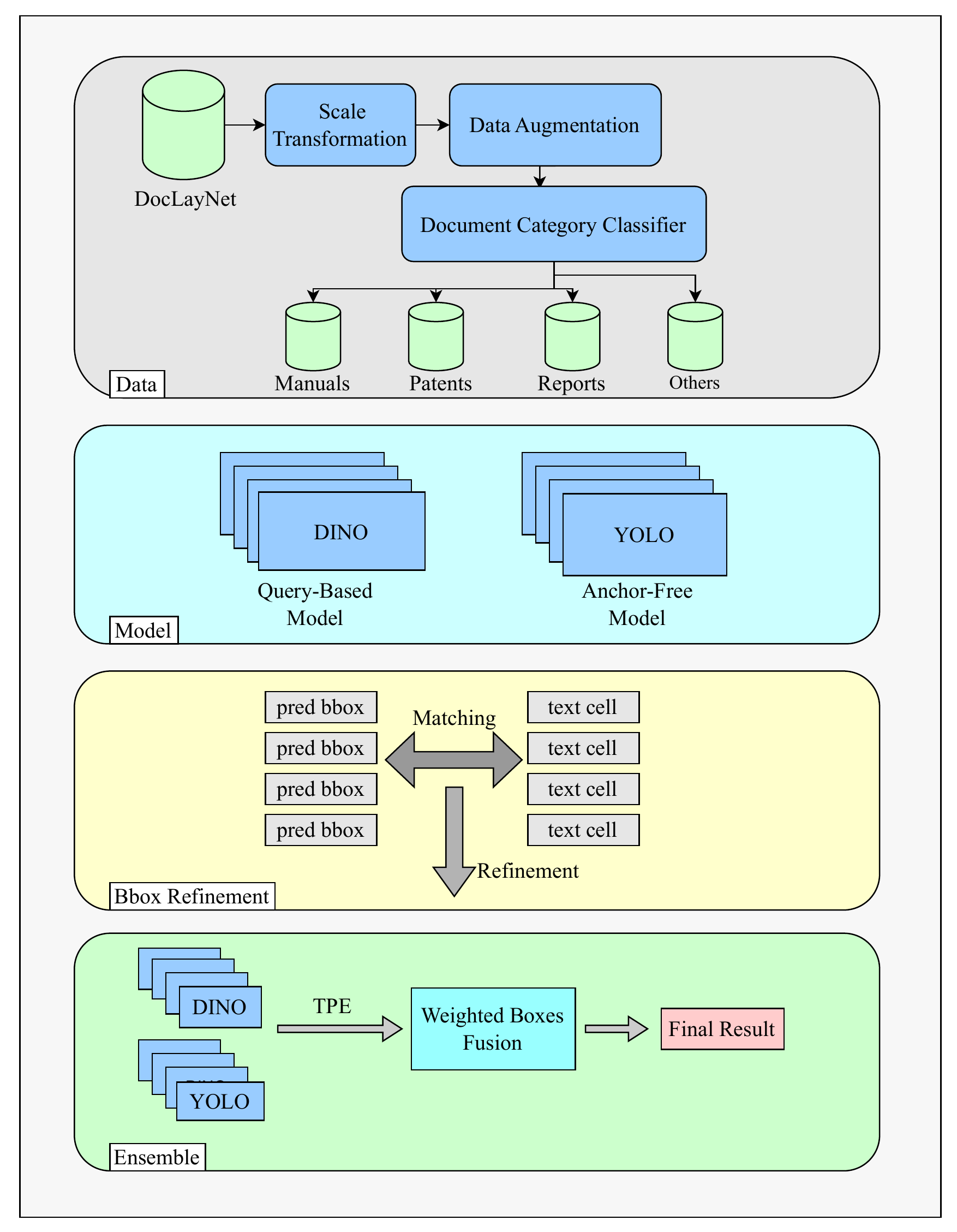} 
\caption{WeLayout: WeChat Layout Analysis System} 
\label{WeLayout} 
\end{figure}

Our team's ensemble of DINO\cite{zhang2022dino} and YOLO\footnote{https://github.com/ultralytics/ultralytics} models, with Weighted Box Fusion\cite{solovyev2021weighted}, achieved a mAP of 70.0 on the competition dataset, securing first place on the leaderboard and significantly outperforming the baseline. Our primary focus was on optimizing various aspects of the task, including the datasets, models, bounding box refinement, and model ensemble. We conducted separate data training for each document category and utilized query-based and anchor-free models to achieve optimal performance. Additionally, we developed an algorithm for cell matching, which significantly improved our performance, and employed Bayesian optimization using a Tree-Structured Parzen Estimator\cite{bergstra2011algorithms} to find near-optimal weights and IoU thresholds for our model ensemble. We will discuss each component in detail in the subsequent sections. The flow chart of our system is shown in Fig. \ref{WeLayout}.

\section{Dataset}

In this section, we introduce the details of training data and data augmentation methods we use in this task.

\subsection{DocLayNet}
The DocLayNet\cite{pfitzmann2022doclaynet} dataset has been released by the organizers for the purposes of training, testing, and competition. This comprehensive dataset comprises over 80,000 human-annotated document pages, showcasing a wide variety of layouts. It categorizes layout components into 11 distinct classes, including paragraphs, headings, tables, figures, lists, mathematical formulas, and several others. It is important to note that DocLayNet offers more than just bitmap page samples and COCO bounding-box annotations; it also includes JSON files containing the digital text-cells extracted from the original PDFs.

\subsection{Data Augmentation}


In an effort to enhance model performance, we employed data augmentation by generating synthetic document images for model training, thereby expanding the diversity and size of the training dataset.

More specifically, we employed a two-step pipeline method for data synthesis, consisting of cropping and composition stages. In the cropping stage, in addition to the training images in DocLayNet, we collected document images from IIIT-AR-13K\footnote{http://cvit.iiit.ac.in/usodi/iiitar13k.php} and TNCR\footnote{https://github.com/abdoelsayed2016/TNCR\_Dataset}, which contain an abundance of images featuring pictures, tables, and other objects. Utilizing the corresponding bounding box information and prevalent typesetting rules, we cropped out all objects from the original document images.

During the composition stage, we initially determined the number of columns, ranging from one to five, and subsequently selected a random background image from our assembled image set. To incorporate objects into the background, for \textit{Title}, \textit{Page-header}, \textit{Page-footer}, and \textit{Footnote} layout elements, we added them with different probabilities, respectively. For other layout types, we randomly selected them from the pre-cropped sets and sequentially pasted them onto the background image column by column. Consequently, we generated a synthetic dataset comprising 300,000 images. Some synthesized images are demonstrated in Fig. 2.

\begin{figure}[h] 
\centering 
\includegraphics[width=0.98\textwidth]{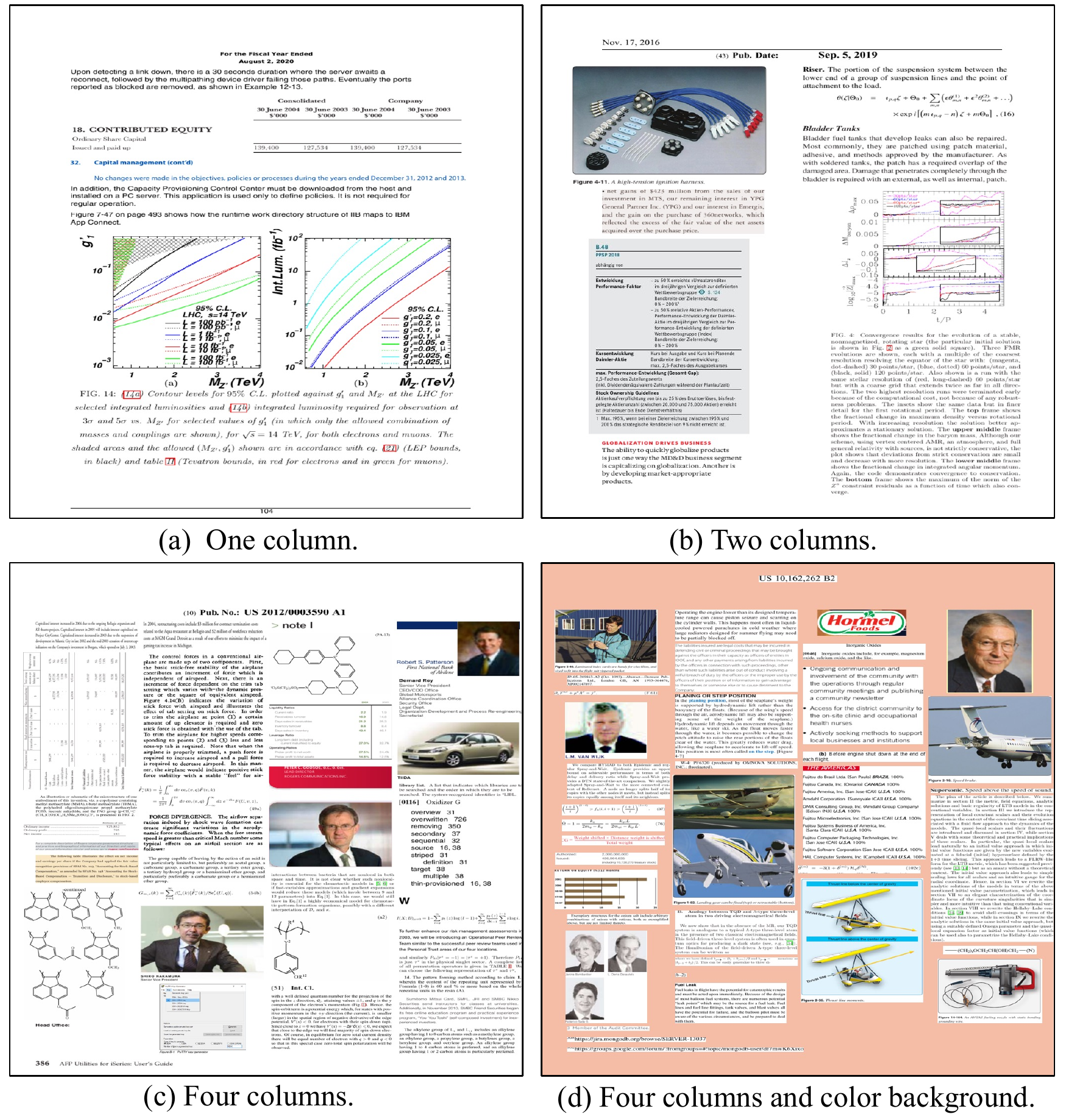} 
\caption{The synthesized images with different layouts.} 
\label{Fig.2} 
\end{figure}

\subsection{Scale Transformation}
In order to better maintain the normal shape of the text, we restored the proportion of the picture based on the scale information in the additional data.

\subsection{Document Category Classification}
The final mAP score is calculated as the mean mAP across four document categories: \textit{reports}, \textit{manuals}, \textit{patents}, and \textit{others}. To improve task-specific learning, it is essential to classify the document images into these categories. It is important to note that the \textit{others} class is not present in the training data; therefore, we train a three-class classification model using ViT\cite{dosovitskiy2020image}.

Additionally, we perform label matching since the document category labels in the training data differ from the \textit{reports}, \textit{manuals}, \textit{patents}, and \textit{others} used in leaderboard score calculation. Specifically, we map the original labels \textit{Scientific Articles}, \textit{Laws and Regulations}, \textit{Government Tenders}, and \textit{Financial Reports} to \textit{reports}. For \textit{manuals} and \textit{patents}, the original labels already include these two classes, so we retain the original labels. After training the document category classifier, we calculate the document category probabilities for the 498 competition data points and select those images with top-1 probability below 0.5 as \textit{others}. Ultimately, we obtain category counts of 292 for \textit{reports}, 114 for \textit{manuals}, 43 for \textit{patents}, and 49 for \textit{others}, which closely aligns with the official split of 293, 114, 48, and 43.

Once the document categories are determined, we conduct separate data training for each document category, which yields promising results. For instance, we observed that the YOLOv8 model trained on all data achieves a mAP of 61.0 on \textit{manuals}, but this score increases to 63.0 when trained exclusively on \textit{manuals}.

\section{Model}
Contemporary object detectors can be broadly categorized into three types: anchor-based models\cite{ren2015faster, liu2016ssd}, anchor-free models\cite{huang2015densebox, redmon2016you, law2018cornernet}, and query-based models\cite{carion2020end, zhu2020deformable}. We have experimented with various models and discovered that both query-based and anchor-free models yield satisfactory detection results for this task.

\subsection{Query-Based Model}

Query-based detectors represent a category of object detection models that leverage learnable queries to examine image features and execute set-based box predictions. Two notable examples of query-based detectors include DETR\cite{carion2020end} and DINO\cite{zhang2022dino}.

DETR (\textbf{DE}tection \textbf{TR}ansformer) is a detection model that employs learnable queries to probe image features derived from the output of Transformer encoders. It utilizes bipartite graph matching to carry out set-based box predictions. DINO (\textbf{D}ETR with \textbf{I}mproved de\textbf{N}oising anch\textbf{O}r boxes) is a cutting-edge, end-to-end object detector that surpasses previous DETR-like models in terms of performance and efficiency. This is achieved through a contrastive approach for denoising training, a mixed query selection method for anchor initialization, and a look-forward-twice scheme for box prediction. In this competition, we opt for the DINO model.

To enhance the detection performance of the DINO model for small text objects, we employ scaled data and carefully design multi-scale augmentations. Furthermore, we incorporate the latest focal modulation network\cite{yang2022focal} as the backbone to boost the model's performance. The results presented in Table \ref{tab:dino} demonstrate a significant improvement over the baseline.

\begin{table}[h]
\centering
\caption{mAP on the combined validation and test sets for DINO. Bold indicates the highest value.}
\setlength\tabcolsep{10pt} 
\begin{tabular}{lc}
\toprule
\textbf{methods} & \textbf{mAP} \\
\midrule
baseline & 76.8 \\
\midrule
DINO & 77.6 \\
\ \  + Augmentation & 78.4 \\
\ \  + Scaled Data & 79.0 \\
\ \  + Focal Network & \textbf{80.7} \\
\bottomrule
\end{tabular}

\label{tab:dino}
\end{table}

\subsection{Anchor-Free Model}
Anchor-free models, a subtype of object detection models, address object detection problems using a per-pixel prediction approach akin to segmentation. This simplification not only streamlines computations but also removes the need for anchor boxes as a hyperparameter. One prominent example of an anchor-free model is FCOS\cite{tian2019fcos}, which serves as the foundation for most recent anchor-free or anchorless deep learning-based object detectors.

In our investigation of existing anchor-free models, we found that YOLOv8 surpasses its counterparts in the given task. As the latest iteration in the YOLO series\cite{redmon2016you, redmon2017yolo9000, redmon2018yolov3, glenn_jocher_2022_7002879, li2022yolov6, li2023yolov6}, YOLOv8 amalgamates numerous features from single-stage detectors, striking an ideal balance between speed and performance. To optimize the task, we meticulously examined preprocessing steps and eliminated superfluous ones, including flip and mosaic augmentation.

Although larger models typically deliver superior results, our findings indicate that performance exceeding a medium-sized model is sufficient for model ensembles post-processing. As such, we trained medium, large, and extra-large YOLOv8 P5 models, as well as an extra-large P6 model. Considering the original image size of 1025x1025, we adjusted the resolution to 1024x1024 and incorporated a 1536x1536 resolution for the P6 model to offer an expanded receptive field. Our experiments showed a significant improvement over the baseline, as illustrated in Table \ref{tab:yolo}.

\begin{table}[h]
\centering
\caption{mAP on the combined validation and test sets for YOLO. Bold indicates the highest value.}
\setlength\tabcolsep{10pt} 
\begin{tabular}{lc}
\toprule
\textbf{methods} & \textbf{mAP} \\
\midrule
baseline & 76.8 \\
\midrule
YOLOv8 medium & 77.3 \\
\ \  + Extra-large & 78.1 \\
\ \  + P6 Model & \textbf{79.9} \\
\bottomrule
\end{tabular}

\label{tab:yolo}
\end{table}

\section{Bounding Box Refinement}
With the release of the additional JSON file by the organizer, we can utilize the text cells' axis information to refine the detection results. It is crucial to determine which cells belong to the detected bounding boxes (bboxs). To address this, we developed an algorithm for cell matching, which significantly improved our performance.

We categorize cell matching into five scenarios and refine the model's predictions in two of them:

\begin{enumerate}
    \item None of the four coordinates (left, top, right, bottom) of the detected bbox are close to any text cells. We skip this case.
    \item One of the four coordinates (left, top, right, bottom) of the detected bbox is close to some text cells. Since the information from a single-edge match is insufficient, we also skip this case.
    \item Two of the four coordinates (left, top, right, bottom) of the detected bbox are close to some text cells. For the other two non-close edges, there are three situations: both are within the text cells, both are outside the text cells, or one is inside while the other is outside the text cells. If two edges are within the text cells and the other two are close to the text cells, it indicates that the detector missed some areas. In this case, we directly replace the predicted results with the most close text cell. If two edges are outside the text cells and the other two are close to the text cells, this occurs when the predicted area contains multiple text cells. We also add the text cells to candidate bboxs for further alignment. For the last situation, we cannot determine the relationship between the predicted area and the text cell, so we skip it.
    \item Three of the four coordinates (left, top, right, bottom) of the detected bbox are close to some text cells. In this case, we carefully assess whether the predicted bbox is within the given text cells. If it is, we directly replace the result with the given text cell axis, as a word cell is a subset of layout elements. This occurs for some layout elements that are long but only one line, which anchor-free models struggle to handle. If the predicted bbox is outside the given text cells, meaning the predicted bbox is a line consisting of several text cells, we add the matched text cell to the candidate bbox for further adjustment.
    \item All four coordinates (left, top, right, bottom) of the detected bbox are close to some text cells. This indicates a good model prediction. However, the detailed axis may not be accurate enough, so we simply replace the prediction axis with the given text cell axis.
\end{enumerate}

Finally, if the detected bbox has not been modified and there are candidate bboxs for it, we choose the closest text cell axis to adjust the predicted result.

It is crucial to emphasize that text cells exclusively contain text; therefore, images and tables do not benefit from this post-processing step. This refinement substantially enhances the performance of DINO and YOLO models, as evidenced by the metrics presented in Table \ref{tab:post_process}:

\begin{table}[h]
\centering
\caption{mAP on the combined validation and test sets for methods without and with post-processing. Bold indicates the highest value, while underline denotes the second highest value.}
\setlength\tabcolsep{10pt} 
\begin{tabular}{lc}
\toprule
\textbf{methods} & \textbf{mAP} \\
\midrule
DINO & 80.7 \\
\ \  + Post-processing & \textbf{89.9}(+9.2) \\
\midrule
YOLO & 79.9 \\
\ \  + Post-processing & \underline{88.5}(+8.6) \\
\bottomrule
\end{tabular}

\label{tab:post_process}
\end{table}

This table demonstrates the significant improvement in model performance when post-processing is applied, with the DINO model achieving the highest mAP value of 89.9, followed by the YOLO model at 88.5.

\section{Ensemble}
\begin{figure}[!h] 
\centering 
\includegraphics[width=0.65\textwidth]{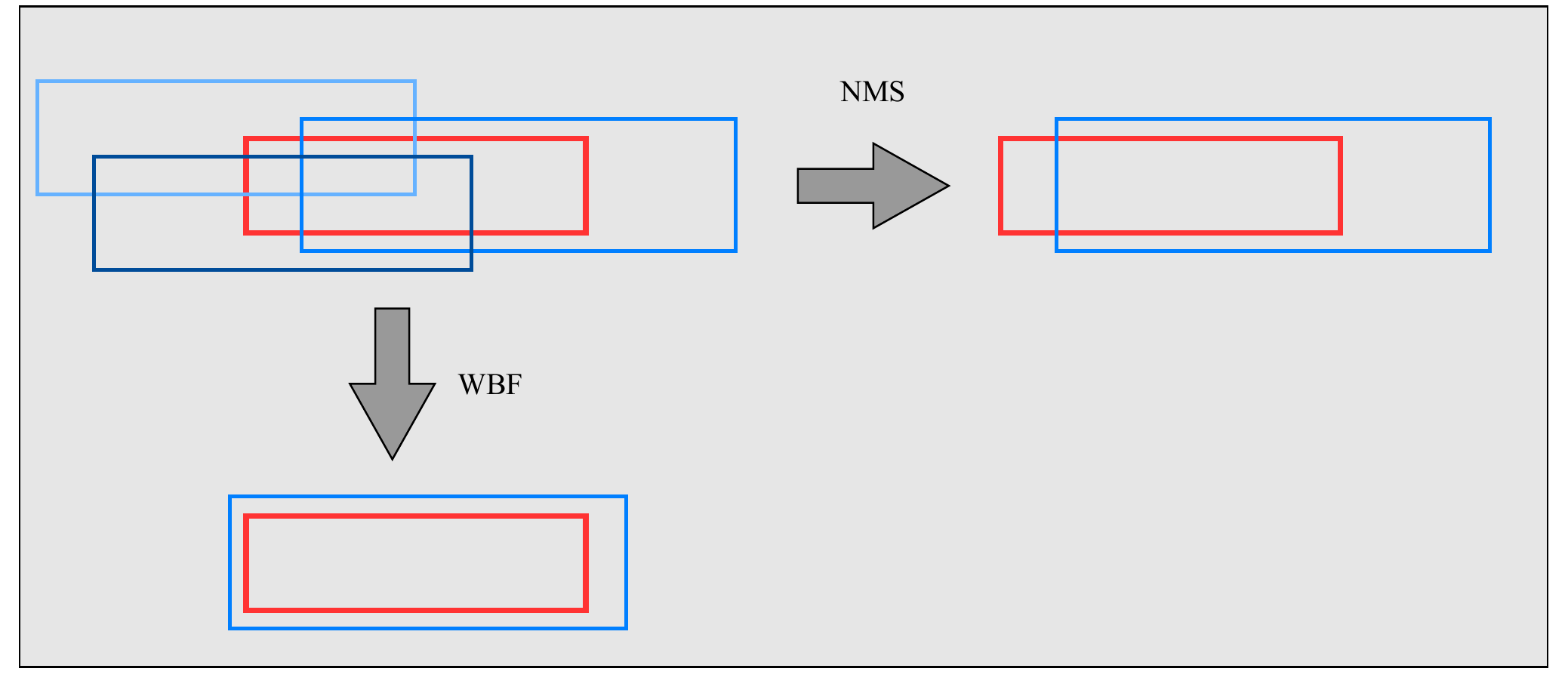} 
\caption{Schematic illustration  of Weighted Boxes Fusion (WBF) vs Non-Maximum Suppression (NMS), blue for DINO and YOLO's predictions, red for ground truth.}\label{WBF} 
\end{figure}

In order to combine the strengths of query-based and anchor-free models, we employ Weighted Boxes Fusion (WBF)\cite{solovyev2021weighted} to merge the predictions from both model types. The specific process is shown in Fig.  \ref{WBF}. Specifically, we integrate DINO and YOLO models, which consist of approximately 10 models, resulting in a vast hyperparameter space. Assuming the weights of all models range from 0 to 10, and the IoU threshold for WBF lies between 0.01 and 0.99, the total hyperparameter space is nearly $10^{12}$. This size is too large to efficiently search for the best result.

To address this issue, we apply Bayesian optimization using a Tree-Structured Parzen Estimator (TPE)\cite{bergstra2011algorithms} to find near-optimal weights and IoU thresholds, as the hyperparameter values are all discrete. Additionally, we leverage Ray\cite{moritz2018ray} for parallel hyperparameter optimization across 64 processes. This approach enables us to determine the optimal weights within a day, making it practical for competition purposes. The results are impressive; for instance, an ensemble of checkpoints with mAP values of (89.9, 89.7, 88.7, 89.0, 87.2, 87.3, 87.5, 88.5, 87.6) achieves a mAP of 91.0 after 2500 search attempts.

\section{Results and Analysis}
In conclusion, we present the results of our work in Table \ref{tab:results}, which showcases the mean average precision (mAP) for all the optimization techniques we proposed. We found that there are differences in the distribution of the test set and the validation set. In order to evaluate the model more comprehensively and obtain better results on the competition set, we combined the test set and the validation set to evaluate the effect of the model.

\begin{table}[h]
\centering
\caption{mAP on the combined validation and test sets for all the optimization techniques we proposed. Bold indicates the highest value.}
\setlength\tabcolsep{10pt} 
\begin{tabular}{lc}
\toprule
\textbf{Methods}             & \textbf{mAP}  \\ 
\midrule
Baseline            & 76.8 \\
\midrule
DINO                & 77.6 \\
\ \  + Augmentation      & 78.4 \\
\ \  + Scaled Data & 79.0 \\
\ \  + Focal Network     & 80.7 \\
\ \  + Post-processing      & 89.9 \\
\midrule
YOLOv8 Medium       & 77.3 \\
\ \  + Extra-large       & 78.1 \\
\ \  + P6 Model          & 79.9 \\
\ \  + Post-processing      & 88.5 \\
\midrule
Weighted Boxes Fusion & \textbf{91.0} \\ 
\bottomrule
\end{tabular}

\label{tab:results}
\end{table}

As observed in Table \ref{tab:results}, the cell matching algorithm employed as a post-processing step significantly improves the final mAP. This suggests that the raw model predictions are close to the ground truth but fail to predict with high precision. The incorporation of additional techniques, such as augmentation, focal network, and the P6 model, further enhances the performance of the AI system. The Weighted Boxes Fusion method achieves the highest mAP, demonstrating its effectiveness in improving the overall accuracy of the model.

\section{Conclusion}

In this paper, we presented our approach to the ICDAR 2023 Competition on Robust Layout Segmentation in Corporate Documents, which involved the development of an ensemble of DINO and YOLO models with Weighted Box Fusion. Our method focused on optimizing various aspects of the task, including dataset preparation, model selection, bounding box refinement, and model ensemble. By conducting separate data training for each document category and utilizing query-based and anchor-free models, we achieved optimal performance. Furthermore, our algorithm for cell matching and the application of Bayesian optimization using a Tree-Structured Parzen Estimator significantly contributed to our success.

Our ensemble achieved a mAP of 70.0 on the competition dataset, securing first place on the leaderboard and significantly outperforming the baseline. This accomplishment demonstrates the effectiveness of our approach in addressing the challenges of document layout segmentation in corporate documents. We believe that our methodology can be further refined and extended to other document analysis tasks, contributing to the advancement of AI-based document processing and information extraction.

%
%
%
%




\printbibliography 

\appendix

\section{Final Leaderboard}
\begin{figure}[!h] 
\centering 
\includegraphics[width=0.9\textwidth]{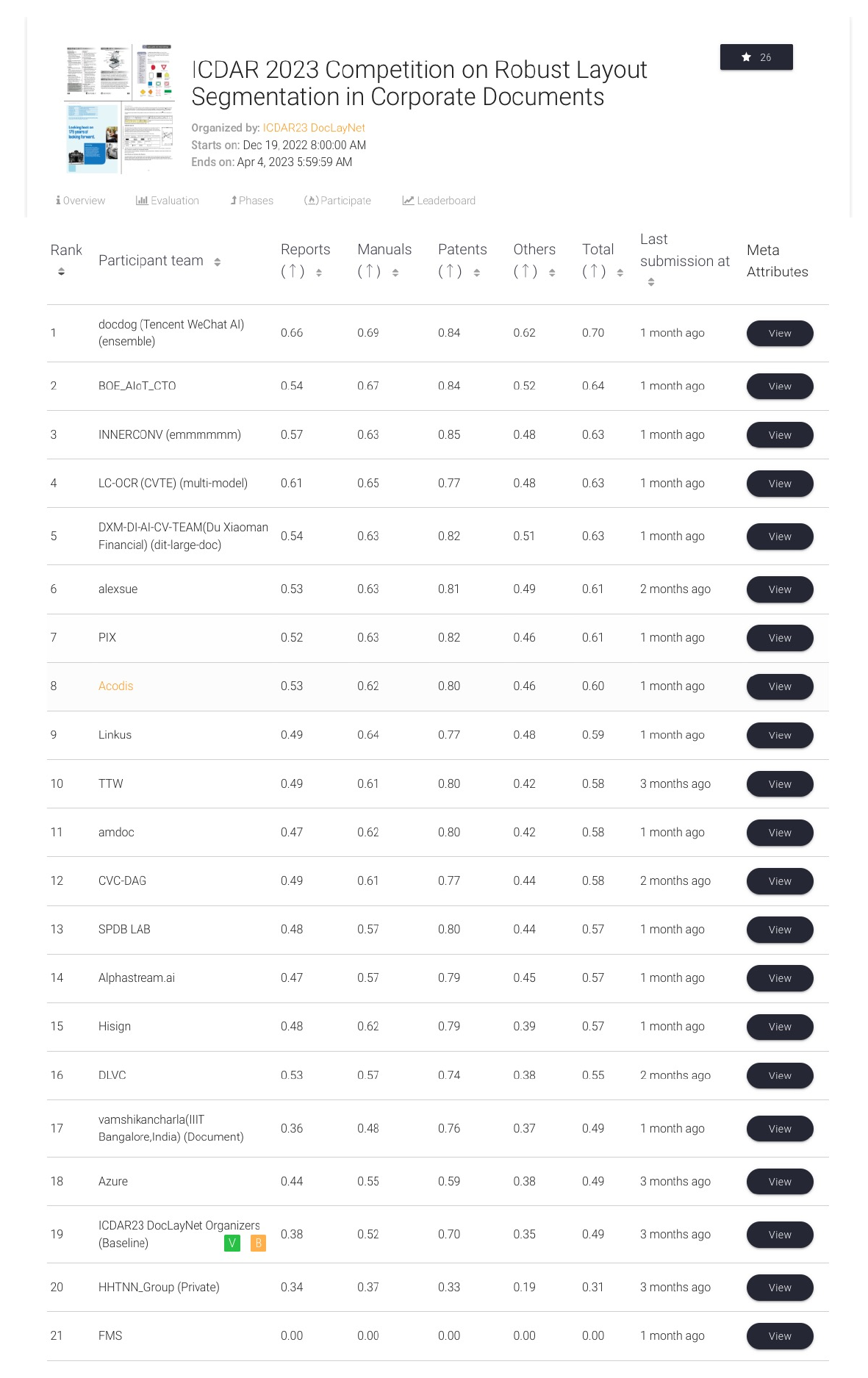} 
\caption{The final leaderboard for the ICDAR23-DocLayNet} 
\label{leaderboard} 
\end{figure}

In Figure \ref{leaderboard}, it is evident that our team has achieved a mean average precision (mAP) of 70.0, which significantly outperforms the competing teams. Furthermore, our approach surpasses the second-place team by 10 mAP in the "others" document category, showcasing our superior generalization capabilities.



\end{document}